\documentclass[letterpaper]{article} %
\usepackage[preprint]{aaai2027}
\usepackage[hyphens]{url}  %
\usepackage{graphicx} %
\usepackage{natbib}  %
\usepackage{caption} %
\usepackage{algorithm}
\usepackage{algorithmic}

\usepackage{newfloat}
\usepackage{listings}
\DeclareCaptionStyle{ruled}{labelfont=normalfont,labelsep=colon,strut=off} %
\floatstyle{ruled}
\newfloat{listing}{tb}{lst}{}
\floatname{listing}{Listing}

\usepackage{booktabs}
\usepackage{amsmath}
\usepackage{amssymb}
\usepackage{multirow}

\nocopyright

\newcommand{\method}{MFlowHOI}
\newcommand{\rvq}{RVQVAE}
\title{\method: Whole-Body Human-Object Interaction Generation\\via Part-Aware MeanFlow}
\author{
    Qiang He\textsuperscript{\rm 1,\rm 2},
    Yonghao Zhang\textsuperscript{\rm 1,\rm 2},
    Yinda Zhang\textsuperscript{\rm 3},
    Chenyu Meng\textsuperscript{\rm 1,\rm 2},
    Liufei Ouyang\textsuperscript{\rm 1,\rm 2},\\
    Haoxiang Hu\textsuperscript{\rm 1,\rm 2},
    Cangjun Gao\textsuperscript{\rm 1,\rm 2},
    Xiaoming Deng\textsuperscript{\rm 1,\rm 2},
    Cuixia Ma\textsuperscript{\rm 1,\rm 2},
    Hongan Wang\textsuperscript{\rm 1,\rm 2}
}
\affiliations{
    \textsuperscript{\rm 1}Institute of Software, Chinese Academy of Sciences\\
    \textsuperscript{\rm 2}University of Chinese Academy of Sciences\\
    \textsuperscript{\rm 3}Google
}

\begin{document}

\maketitle
\begin{abstract}
Whole-body human--object interaction (HOI) generation requires coordinating
body and finger motion, object trajectories, and time-varying contacts.
Existing methods typically rely on iterative diffusion or test-time
optimization, resulting in high inference costs. 
Direct one-step generation is highly desired in downstream applications due to efficient inference. However, it is challenging because these HOI components differ substantially in structure and scale. 
We present \method{}, an efficient one-step framework for
whole-body HOI generation. A Contact-Aware Multipart \rvq{} organizes body,
hand, object, and contact signals into a structured latent space, with a shared
decoder preserving their global coordination. An ST-DiT then models temporal
and cross-part dependencies in this space and generates a complete interaction
in a single MeanFlow evaluation. Predicted object-local contact
correspondences further enable geometric refinement and
interaction-preserving transfer to photorealistic 3D Gaussian avatars with
different body proportions. 
Experiments demonstrate that \method{} outperforms existing base
generators in motion quality and text alignment while substantially reducing
inference time.
\end{abstract}

\section{Introduction}
\label{sec:intro}
Human--object interaction (HOI) generation supports applications in virtual
reality, character animation, embodied simulation, and robotics. It requires
the body, articulated hands, and object to move coherently while maintaining
plausible spatial relations over time. This is particularly challenging for
dexterous interactions: small relative-pose errors can cause unreachable
grasps, floating contacts, or severe penetration despite otherwise natural
motion.
\begin{figure}[t]
  \centering
  \includegraphics[width=\columnwidth]{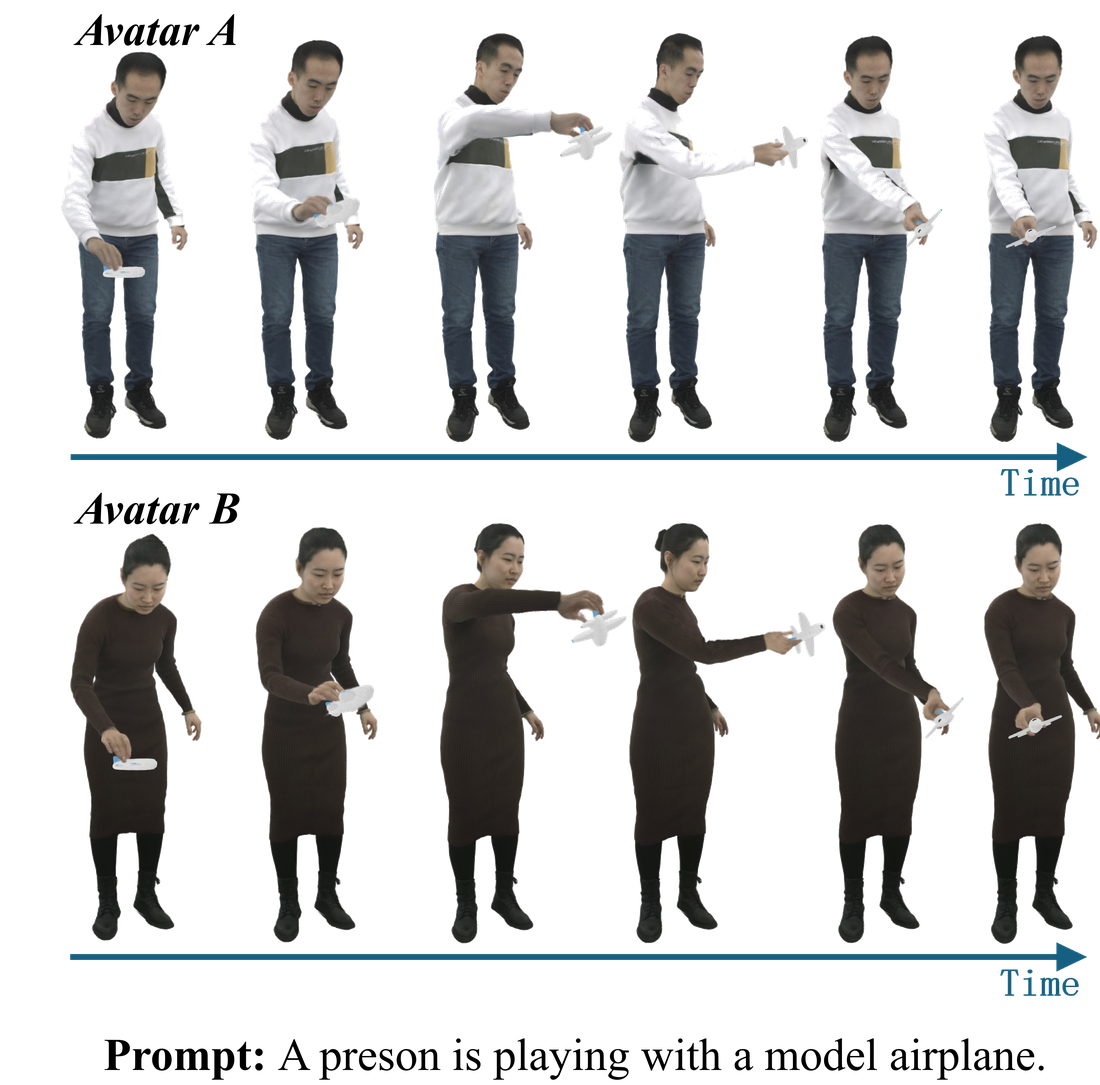}
  \caption{\method{} jointly generates body and finger motions, object
trajectories, and contacts in a single transport pass, enabling refinement and retargeting to 3D Gaussian avatars with diverse body proportions.}
  \label{fig:teaser}
\end{figure}

Existing methods capture complementary aspects of HOI. Hand--object approaches
model articulated hands and local contact geometry
~\citep{christen2022d,liu2023contactgen,cha2024text2hoi,liu2024geneoh},
but omit the full-body motions to drive the hands. Whole-body methods coordinate human and
object trajectories
~\citep{ghosh2023imos,li2023object,li2024controllable,peng2025hoi}, yet often use
simplified hand representations. Recent methods jointly model body and finger
motion
~\citep{zhang2025diffgrasp,ron2025hoidini,pi2026coda,pratikaki2026dex2hoi},
but typically rely on iterative diffusion, geometric guidance, or test-time
optimization to maintain contact
~\citep{li2024controllable,peng2025hoi,zhang2025diffgrasp,
karunratanakul2024optimizing,ron2025hoidini,pi2026coda}. Although high-quality results can be achieved, their high inference cost
limits interactive use.
Recently, MeanFlow~\citep{gengmean} enables one-step generation by learning an
interval-averaged velocity field and has shown promising results in human
motion synthesis~\citep{yang2025flowerdance,geng2025armflow}. However, it is nontrivial to apply it to HOI because body motion, finger articulation,
object trajectories, and sparse contacts between human and objects have different structures and scales. Directly encoding them jointly can suppress subtle hand and contact dynamics, whereas modeling them independently weakens the quality of interaction coordination. In our
experiments, we find that directly applying one-step MeanFlow to existing HOI representations leads to degraded motion and unstable contact.

In this paper, we propose \method{}, a one-step framework for contact-aware
whole-body HOI generation. Its Contact-Aware Multipart RVQ-VAE decomposes an
interaction into functional body parts, object motion, contact probabilities,
and object-local contact positions. Stream-specific temporal encoders capture
fine-grained dynamics, while a shared decoder preserves global coordination,
forming a structured time--part latent space.
In this latent space, a conditional Spatio-Temporal Diffusion Transformer
(ST-DiT) learns the MeanFlow velocity field. It models temporal dependencies
within each stream and cross-part interactions among the body, hands, object,
and contacts. A single evaluation generates a complete interaction latent,
which is decoded into whole-body and finger motion, object trajectories, and
contact cues. The resulting contact correspondences also enable geometric
refinement and interaction-preserving transfer to photorealistic 3D Gaussian
avatars with different body proportions.
Experiments on GRAB demonstrate improved distributional quality and
text--motion alignment with substantially faster inference than existing HOI
generators. Ablations confirm the importance of the multipart latent space and its temporal and cross-part modeling.

The main contribution of our method can be summarized as follows. 
1) We present a new one-step framework that jointly generates coordinated whole-body and finger motion, object trajectories, and time-varying contact cues with a single network. 
2) To enable stable conditional MeanFlow generation, we design a temporal multipart--based HOI latent representation to model temporal dependencies between different body parts and objects, and introduce a spatio-temporal diffusion transformer to balance heterogeneous HOI components.
3) We predict contact correspondences between object and body parts that support geometric refinement and interaction-preserving transfer to photorealistic 3D Gaussian avatars with different body proportions.

\section{Related Work}
\label{sec:related}
\paragraph{Efficient and Controllable Motion Generation.}
Human motion generation has been extensively studied under various conditions,
including text, music, action labels, keyframes, and joint trajectories.
Diffusion-based methods have shown strong generative performance: MDM~\citep{tevethuman}
introduced a general framework for text- and action-conditioned motion
generation, while OmniControl~\citep{xie2024omnicontrol} and TLControl~\citep{wan2024tlcontrol} extended
diffusion models with fine-grained spatial and trajectory constraints. DNO~\citep{karunratanakul2024optimizing}. Further treats a pretrained
diffusion model as a motion prior and optimizes its initial noise to satisfy
test-time objectives. Despite their flexibility,
these approaches require iterative denoising or additional optimization during
inference. Flow-based models offer a more efficient alternative: Motion Flow
Matching~\citep{hu2023motion} generates motion through continuous-time transport, and FlowMotion~\citep{cuba2025flowmotion} improves temporal coherence using
target-predictive conditional flow matching.
Recently, FlowerDance~\citep{yang2025flowerdance} and ARMFlow~\citep{geng2025armflow} explored MeanFlow~\citep{gengmean} for one-step dance and
human-reaction generation.
However, efficient one-step generation remains underexplored for
human--object interactions, where human motion must be coordinated with object
motion and time-varying contact.

\paragraph{Hand--Object Interaction Generation.}
Hand--object interaction generation focuses on synthesizing plausible hand
poses or motion with respect to an object. Early studies commonly generate
static grasps using object geometry, contact maps, or interaction
priors. Recent methods extend this setting to dynamic interactions:
ContactGen~\citep{liu2023contactgen} represents contacts through object-surface locations, hand parts,
and contact directions; D-Grasp~\citep{christen2022d} uses
physics-based control to synthesize dynamic grasping behaviors; Text2HOI~\citep{cha2024text2hoi} predicts a contact map before generating
hand--object motion; and GeneOH~\citep{liu2024geneoh} 
Diffusion
adopts a
contact-centric representation for interaction denoising. 
In addition, several recent studies seek up better cross-modality consistency ~\citep{li2025latenthoi,huang2025hoigpt,christen2024diffh2o} , geometric or physical compliance ~\citep{zhang2021manipnet,taheri2024grip,zhang2024artigrasp,zheng2023cams} and interaction recovering~\citep{zhou2022toch}.
These methods capture detailed hand articulation and local contact geometry, but generally do not model the complete human
kinematic chain. Consequently, a locally plausible hand pose or trajectory
does not necessarily correspond to natural and reachable motion of the arms,
torso, and root.

\paragraph{Human--Object Interaction Generation.}
Human--object interaction generation studies the coupled motion of a person
and a movable object.
Recently, many works focus on full-body generation~\citep{taheri2022goal,wu2022saga,wu2024human,li2024task,kwon2026graspdiffusion} and physically plausible interactions~\citep{braun2024physically,di2026pamotion}. Furthermore, 
IMoS~\citep{ghosh2023imos} generates intent-driven full-body interactions with
objects, while OMOMO~\citep{li2023object} predicts hand locations from a
given object trajectory and uses them to condition body-motion synthesis. CHOIS~\citep{li2024controllable} jointly generates human and object motion from
language, initial states, and sparse object waypoints, using sampling-time
guidance to improve contacts. HOI-Diff~\citep{peng2025hoi} employs separate
human and object diffusion branches together with affordance modeling and
classifier guidance. These methods primarily represent
the body and object while omitting fully articulated finger motion. More recent
approaches model finer hand articulation: DiffGrasp~\citep{zhang2025diffgrasp} generates body and
hand motion conditioned on an object trajectory and applies reconstruction
guidance during diffusion sampling; HOIDiNi~\citep{ron2025hoidini} and
CoDA~\citep{pi2026coda} coordinate body, hand, and object generation through staged diffusion or
noise optimization; and Dex2HOI~\citep{pratikaki2026dex2hoi} avoids
external noise optimization but retains iterative diffusion
sampling.
Therefore, simultaneously modeling body
motion, articulated hands, object motion, and temporal contact remains
challenging, particularly under efficient inference.

\begin{figure*}[t]
  \centering
  \includegraphics[width=\textwidth]{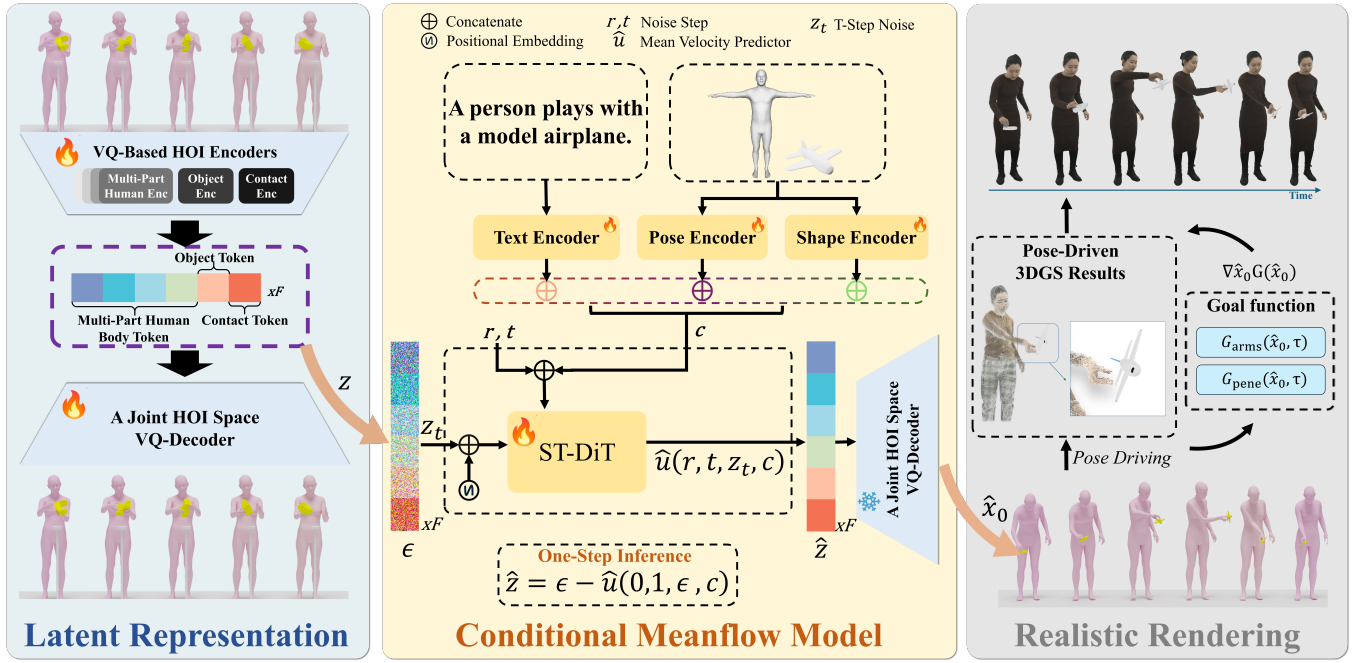}
  \caption{\method{} encodes part-level motion, object motion, object-body part 
  contacts, and contact states into a structured latent. Conditional MeanFlow
  transports noise to a complete HOI latent in one ST-DiT transport evaluation; the frozen
  decoder recovers motion for direct use, optional refinement, or 3DGS avatar
  deployment.}
  \label{fig:overview}
\end{figure*}
\section{Method}
\label{sec:method}
Given the initial poses of a human and an object, the object geometry, and a
textual instruction, our goal is to generate a complete 3D human--object
interaction (HOI) sequence, including whole-body human motion, articulated
finger motion, and object motion over time. As shown in
Figure~\ref{fig:overview}, \method{} consists of three main modules. In
{HOI Latent Representation}, we design a contact-aware multipart
RVQ-VAE to encode multipart human motion, object motion, and contact information into a temporal multipart-based HOI latent representation. In {Conditional MeanFlow}, we employ an ST-DiT with spatial--temporal and cross-part attention to generate the HOI latent sequence in a single transport evaluation, conditioned on the textual instruction, initial poses, and object geometry. In {Realistic Rendering}, we develop a pose-driven 3D Gaussian Splatting framework to produce high-fidelity renderings of the generated 3D HOI sequences.

\subsection{Data Representation}
\paragraph{Human and Object Motion Representation.}
Following DiffGrasp~\cite{zhang2025diffgrasp}, we use SMPL-X to represent human
motion. A human motion sequence is denoted by
$\mathbf{H}\in\mathbb{R}^{T\times D_h}$, where each state is represented as
$\mathbf{H}^{(t)}=\{\mathbf{r}^{(t)},\Delta\mathbf{d}^{(t)},
\boldsymbol{\theta}^{(t)}\}$. Here,
$\mathbf{r}^{(t)}\in\mathbb{R}^{6}$ is the global root orientation,
$\boldsymbol{\theta}^{(t)}\in\mathbb{R}^{52\times6}$ contains the local
rotations of 52 full-body joints, and $\Delta\mathbf{d}^{(t)}$ is the root
translation relative to the previous frame. The initial translation
$\mathbf{d}^{(1)}\in\mathbb{R}^{3}$ is represented separately.
Similarly, an object motion sequence is denoted by
$\mathbf{O}\in\mathbb{R}^{T\times D_o}$, where
$\mathbf{O}^{(t)}=\{\mathbf{R}^{(t)},\Delta\mathbf{o}^{(t)}\}$.
Here, $\mathbf{R}^{(t)}\in\mathbb{R}^{6}$ is the global rotation and
$\Delta\mathbf{o}^{(t)}$ is the translation relative to the previous frame.
The initial object translation $\mathbf{o}^{(1)}\in\mathbb{R}^{3}$ is
represented separately. The human and object shapes are represented by the
SMPL-X shape parameters $\boldsymbol{\beta}$ and the BPS descriptor
$\mathbf{B}\in\mathbb{R}^{K\times3}$, respectively. Overall, the human and
object representations are
$(\mathbf{H},\mathbf{d}^{(1)},\boldsymbol{\beta})$ and
$(\mathbf{O},\mathbf{o}^{(1)},\mathbf{B})$, respectively.

\paragraph{Interaction Representation.}
Following HOIDiNi and CoDA~\cite{ron2025hoidini,pi2026coda}, we represent
human--object interactions using human-centric contact cues expressed in the
object coordinate frame. Denote $\mathcal{M} = \{m_i\}_{i=1}^{N}$ to be a set of predefined body markers. At frame $t$, each marker $m_i$ is associated
with a contact probability $\gamma_i^{(t)}\in[0,1]$ to indicate whether the marker is in contact with the object, and an object-relative contact
position $\mathbf{p}_i^{(t)} \in \mathbb{R}^{3}$ to specify the contact position in object coordinate frame. The frame-level
interaction is represented as
$\mathcal{I}^{(t)} =
\{\gamma_i^{(t)}, \mathbf{p}_i^{(t)})\}_{i=1}^{N}$, and the interaction
sequence is denoted by
$\mathcal{I} = \{\mathcal{I}^{(t)}\}_{t=1}^{T}$.

\paragraph{Text Description.}
Following IMoS~\cite{ghosh2023imos} and Text2HOI~\cite{cha2024text2hoi},
we describe each interaction using the template
``A person $\langle\text{action}\rangle$ the $\langle\text{object}\rangle$.''

\subsection{HOI Latent Representation}
\label{sec:hoi_latent_representation}
We design a multipart residual vector-quantized variational autoencoder
(RVQ-VAE)~\cite{lee2022autoregressive} to learn compact
discrete representations of human--object interactions. The human motion,
object motion, and interaction cues are separately encoded and quantized,
and then jointly reconstructed by a shared decoder.

\paragraph{Multipart Encoding.}
We divide the human motion $\mathbf H$ into semantic streams $\mathcal S_h$ of eight typical body parts, corresponding to the root, head, left and right hands, left and right legs,
and left and right feet. The object motion $\mathbf O$ forms an additional
stream, while the interaction representation $\mathcal I$ is separated into
a contact-probability stream $\boldsymbol{\Gamma}
=\{\gamma_i^{(t)}\}_{t=1,i=1}^{T,N}$ and an object-relative
contact-position stream $\mathbf P
=\{\mathbf p_i^{(t)}\}_{t=1,i=1}^{T,N}$.
The complete stream set is
$\mathcal S=\mathcal S_h\cup\{\mathrm{O},\mathrm{\Gamma},\mathrm{P}\}$.
For each stream $s\in\mathcal S$, a stream-specific temporal encoder maps
its input $\mathbf X_s$ to a continuous latent sequence
$\mathbf Z_s=E_s(\mathbf X_s)\in\mathbb R^{T'\times d}$, where $T'$ is the
downsampled sequence length and $d$ is the latent dimension.

\paragraph{Residual Vector Quantization.}
Each latent vector $\mathbf z_s^{(\tau)}$ is quantized using an
$L$-level residual codebook
$\mathcal E_s=\{\mathcal E_{s,l}\}_{l=1}^{L}$. The quantized vector is
obtained by replacing $\mathbf z_s^{(\tau)}$ with the sum of its selected
code entries:
\begin{equation}
    \widehat{\mathbf z}_s^{(\tau)}
    =
    Q_s\!\left(\mathbf z_s^{(\tau)}\right)
    =
    \sum_{l=1}^{L}\mathbf e_{s,l,k_l},
    \qquad
    \mathbf e_{s,l,k_l}\in\mathcal E_{s,l}.
    \label{eq:residual_quantization}
\end{equation}
Applying the quantizer over time produces the discrete latent sequence
$\widehat{\mathbf Z}_s$.

\paragraph{Conditional Joint Decoding.}
We encode the initial human--object state as
$\mathbf c_{\mathrm{init}}
=E_{\mathrm{init}}(\mathbf H^{(1)},\mathbf O^{(1)},
\mathbf d^{(1)},\mathbf o^{(1)})$
and the human and object shapes as
$\mathbf c_{\mathrm{shape}}
=E_{\mathrm{shape}}(\boldsymbol{\beta},\mathbf B)$.
All quantized streams are concatenated as
$\widehat{\mathbf Z}
=\bigoplus_{s\in\mathcal S}\widehat{\mathbf Z}_s$
and jointly decoded by
$
    (
        \widehat{\mathbf H},
        \widehat{\mathbf O},
        \widehat{\mathcal I}
   )
    =
    \mathcal D\!(
        \widehat{\mathbf Z},
        \mathbf c_{\mathrm{init}},
        \mathbf c_{\mathrm{shape}}
    ),
    \label{eq:conditional_hoi_decoding}
$
where $\oplus$ denotes feature-wise concatenation. The shared decoder captures
cross-stream dependencies and jointly reconstructs coherent human motion,
object motion, and interaction cues.

\paragraph{Codec Training.}
We train the codec using
\[
    \mathcal L_{\mathrm{RVQ}}
    =
    \sum_{s\in\mathcal S}
    \lambda_s \mathcal L_{\mathrm{rec}}^{s}
    +
    \lambda_{\mathrm{code}} \mathcal L_{\mathrm{code}}
    +
    \lambda_{\mathrm{commit}} \mathcal L_{\mathrm{commit}},
\]
where $\mathcal L_{\mathrm{rec}}^{s}$ is the Smooth L1 loss for continuous
streams and binary cross-entropy for the contact-state stream,
$\mathcal L_{\mathrm{code}}$ updates the codebook embeddings toward the
encoder outputs, and $\mathcal L_{\mathrm{commit}}$ encourages the encoder
outputs to remain close to their residual-quantized representations.

\subsection{Conditional MeanFlow in the HOI Latent Space}
\label{sec:conditional_meanflow}
After training the Contact-Aware Multipart RVQ-VAE, we freeze its encoders and
decoder. Instead of generating discrete RVQ indices, we model the continuous
pre-quantization latents
$\mathbf z_0=\operatorname{Stack}_{s\in\mathcal S}[\mathbf Z_s]
\in\mathbb R^{T'\times|\mathcal S|\times d}$, which form a structured
time--part token grid.
Given a textual instruction $\mathbf y$, we construct the complete generation
condition as
$\mathbf c=F_{\mathrm{cond}}(\mathbf c_{\mathrm{init}},
\mathbf c_{\mathrm{shape}},E_{\mathrm{text}}(\mathbf y))$,
where $E_{\mathrm{text}}$ is a frozen CLIP text encoder. Thus, $\mathbf c$
jointly specifies the initial human--object state, human and object shapes, and
textual instruction.

\paragraph{Conditional MeanFlow Objective.}
Given a clean latent $\mathbf z_0$ and Gaussian noise
$\boldsymbol\epsilon\sim\mathcal N(\mathbf 0,\mathbf I)$, we construct the
linear path $\mathbf z_t=(1-t)\mathbf z_0+t\boldsymbol\epsilon$,$\mathbf v_{\mathrm{gt}}=\boldsymbol\epsilon-\mathbf z_0$.
    
For an interval $[r,t]$, where $0\le r\le t\le1$, MeanFlow models the average
velocity
\begin{equation}
    \mathbf u(\mathbf z_t,r,t\mid\mathbf c)
    =
    \frac{1}{t-r}
    \int_r^t
    \mathbf v(\mathbf z_\tau,\tau\mid\mathbf c)\,d\tau
    \label{eq:meanflow_average}
\end{equation}
Unlike an instantaneous velocity field, $\mathbf u$ directly transports
between interval endpoints and therefore permits a full-interval update.
The average velocity satisfies the MeanFlow identity
\begin{equation}
    \mathbf u+(t-r)\frac{d\mathbf u}{dt}=\mathbf v,
    \qquad
    \frac{d\mathbf u}{dt}
    =
    \partial_{\mathbf z}\mathbf u\,\mathbf v
    +\partial_t\mathbf u.
    \label{eq:meanflow_identity}
\end{equation}
We parameterize it as
$\widehat{\mathbf u}_{\theta}
=\operatorname{ST\text{-}DiT}_{\theta}(\mathbf z_t,r,t,\mathbf c)$.
Its total derivative is efficiently evaluated using the Jacobian--vector
product
$\dot{\widehat{\mathbf u}}_{\theta}
=\operatorname{JVP}_{(\mathbf z,r,t)}
[\widehat{\mathbf u}_{\theta};
(\mathbf v_{\mathrm{gt}},0,1)]$.
We then construct the detached target
$\mathbf u_{\mathrm{tgt}}
=\operatorname{sg}[
\mathbf v_{\mathrm{gt}}
-(t-r)\dot{\widehat{\mathbf u}}_{\theta}]$,
where $\operatorname{sg}[\cdot]$ denotes stop-gradient. This construction
avoids higher-order backpropagation.
To balance heterogeneous motion and contact streams, we define
$\ell_s=\|\widehat{\mathbf u}_{\theta,s}
-\mathbf u_{\mathrm{tgt},s}\|_2^2$ for each $s\in\mathcal S$ and minimize
\begin{equation}
    \mathcal L_{\mathrm{MF}}
    =
    \mathbb E\!\left[
        \frac{1}{|\mathcal S|}
        \sum_{s\in\mathcal S}
        \operatorname{sg}\!\left[
            (\ell_s+\delta)^{-\alpha}
        \right]\ell_s
    \right].
    \label{eq:meanflow_loss}
\end{equation}
We use $\delta=10^{-3}$ and $\alpha=0.5$. During training, we sample and sort
two time values to ensure $r\le t$, and set $r=t$ for half of each batch to
stabilize the instantaneous-flow boundary.

\paragraph{Spatio-Temporal Attention Modeling.}
ST-DiT operates on the $T'\times|\mathcal S|$ time--part token grid. Temporal
attention models the trajectory of each human, object, and contact stream,
while cross-part attention coordinates all streams at each latent time step.
Together, they jointly capture whole-body dynamics, articulated hand motion,
object motion, contact probabilities, and object-local contact positions.
Temporal, part, and modality embeddings preserve the grid structure. The
condition $\mathbf c$ is injected through part-specific adaptive normalization,
allowing each stream to selectively access the textual instruction, initial
human--object state, and shape information. In particular, the contact streams
provide explicit interaction cues that connect human and object motion.

\paragraph{One-Step HOI Generation.}
At inference time, we sample
$\boldsymbol\epsilon\sim\mathcal N(\mathbf 0,\mathbf I)$ and perform a single
ST-DiT evaluation over the full interval:
$
    \widehat{\mathbf z}_0
    =
    \boldsymbol\epsilon
    -
    \widehat{\mathbf u}_{\theta}
    (\boldsymbol\epsilon,0,1,\mathbf c).
    \label{eq:meanflow_one_step}
$
We unpack the generated time--part grid as
$\{\widetilde{\mathbf Z}_s\}_{s\in\mathcal S}
=\operatorname{Unstack}(\widehat{\mathbf z}_0)$ and reconstruct the complete
interaction using the frozen decoder:
$
    (
        \overline{\mathbf H},
        \overline{\mathbf O},
        \widehat{\mathcal I}
    )
    =
    \mathcal D \!(
        \bigoplus_{s\in\mathcal S}\widetilde{\mathbf Z}_s,
        \mathbf c_{\mathrm{init}},
        \mathbf c_{\mathrm{shape}}
    ).
$
Finally, we restore the prescribed initial poses and integrate the human and
object translation increments from $\mathbf d^{(1)}$ and $\mathbf o^{(1)}$.
The base model therefore generates a complete HOI sequence with one ST-DiT
transport evaluation, without iterative integration or projection onto the
residual codebooks.

\subsection{Optional Contact-Guided Refinement}
\label{sec:interaction_refinement}

Our base model directly generates the final sequence without post-processing.
Nevertheless, its predicted interaction representation establishes explicit
correspondences between human markers and object-local contact points, enabling
two optional refinement strategies. Their computational costs are not included
in the reported one-step generation latency.

\paragraph{Object-Only Refinement.}
We fix the generated human motion and object rotation, and optimize only the
object translation. Confident object-body part contacts are projected onto the
object surface and transformed into world coordinates. We then estimate 
bounded, temporally smooth translation offsets that robustly reduce their
discrepancies from the corresponding human markers. The initial object state
is preserved, and MeanFlow is not rerun.

\paragraph{Upper-Body Noise Optimization.}
Alternatively, we fix the object trajectory and predicted contact targets, and
adapt Diffusion Noise Optimization (DNO) to our differentiable one-step
generator. The input noise is optimized using contact, penetration, motion
smoothness, and scene-consistency objectives. An output projection preserves
the human root and lower body, as well as the object trajectory and contact
predictions, such that only the torso, arms, and articulated hands can change.
Each optimization iteration requires only one ST-DiT evaluation and one frozen
decoder pass.
Full objectives and implementation details for optimization strategies are provided in the supplementary material.

\subsection{Pose-Driven 3DGS Realistic Rendering}
\label{sec:realistic_rendering}
\begin{figure}[t]
  \centering
  \includegraphics[width=\columnwidth]{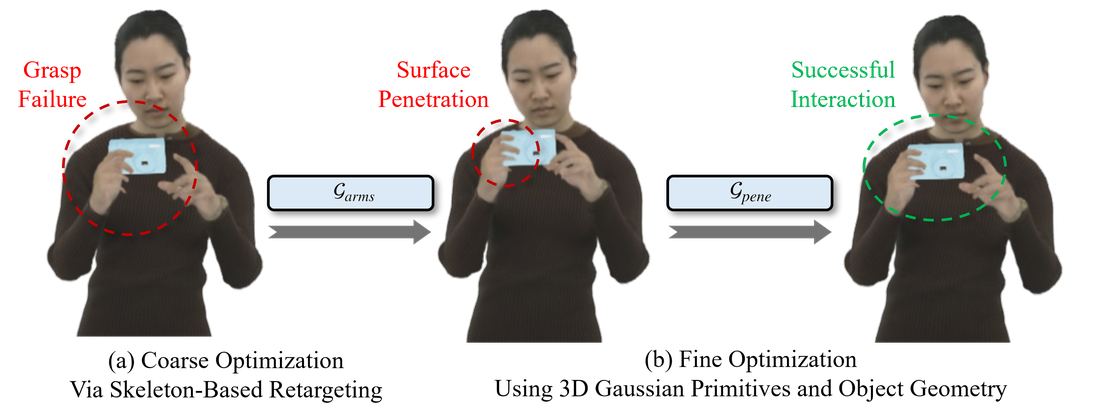}
  \caption{Direct pose transfer across different body shapes may cause grasp failure.
Our coarse arm retargeting first restores reachability, after which fine
hand-Gaussian SDF optimization suppresses local hand--object penetration while
preserving the generated object trajectory.}
  \label{fig:avatar_retargeting}
\end{figure}

To deploy the generated HOI sequences on photorealistic avatars, we develop an
interaction-preserving pose-driven 3D Gaussian Splatting (3DGS) framework.
Directly transferring generated joint rotations to an avatar with different
body proportions changes the global wrist positions, potentially causing grasp
failure and hand--object penetration. As illustrated in
Figure~\ref{fig:avatar_retargeting}, we address these issues through coarse-to-fine
pose adaptation while keeping the generated root and object trajectories fixed.
Let $\widehat{\mathbf x}_0$ denote the generated human pose sequence after
initial retargeting to the target avatar, and let $\boldsymbol\tau$ contain the
source arm-joint targets and fixed object trajectory. We define an arm-alignment objective for restoring reachability and a penetration objective over Gaussians of the hand part:
\begin{equation}
\begin{aligned}
    \mathcal G_{\mathrm{arms}}
    (\widehat{\mathbf x}_0,\boldsymbol\tau)
    &=
    \sum_t
    \left\|
        \mathbf J^{\mathrm{tar}}_{t,\mathrm{arm}}
        (\widehat{\mathbf x}_0)
        -
        \mathbf J^{\mathrm{src}}_{t,\mathrm{arm}}
        (\boldsymbol\tau)
    \right\|_2^2,\\
    \mathcal G_{\mathrm{pene}}
    (\widehat{\mathbf x}_0,\boldsymbol\tau)
    &=
    \sum_t\sum_{g\in\mathcal H}
    \left[
        -\operatorname{SDF}_{O_t(\boldsymbol\tau)}
        \left(
            \boldsymbol\mu_{g,t}(\widehat{\mathbf x}_0)
        \right)
    \right]_+^2,
\end{aligned}
\label{eq:3dgs_guidance}
\end{equation}
where $\mathcal H$ denotes the set of Gaussians of the hand part, and
$\boldsymbol\mu_{g,t}$ is the posed mean of Gaussian $g$. Since the 3DGS
pose-driving function is differentiable, both objectives can be directly
backpropagated to $\widehat{\mathbf x}_0$.
As shown in Figure~\ref{fig:avatar_retargeting}, we apply the two objectives
sequentially. First, optimize the upper-body pose using
$\mathcal G_{\mathrm{arms}}$ to restore reachability across different body
shapes. Starting from the arm-aligned result, then optimize the articulated
hand pose using $\mathcal G_{\mathrm{pene}}$ to suppress local hand--object
penetration. Root motion, lower-body pose, and generated object trajectory remain fixed in both stages.

\section{Experiments}
\label{sec:experiments}

\subsection{Evaluations}

\paragraph{Dataset.}
We evaluate on GRAB~\citep{taheri2020grab}, which contains full-body
interaction sequences captured from 10 subjects interacting with 51 rigid
objects. Following prior work~\citep{pi2026coda,ghosh2023imos}, we use subjects
S1--S9 for training and S10 for evaluation. For the cross-shape deployment
study, we apply the generated interactions to drivable 3DGS avatars with
 AvatarReX~\citep{zheng2023avatarrex}.

\paragraph{Evaluation metrics.}
Following CoDA~\citep{pi2026coda}, we use its released text--motion evaluator to assess generation quality. Fr'echet Inception Distance (FID) measures the distributional gap between generated and real motion features, R-Precision (R@3) evaluates text--motion alignment via Top-3 retrieval accuracy, and Diversity quantifies motion-feature variation relative to the ground-truth (GT) value. For physical quality, foot skating (FS) measures horizontal foot-sliding velocity in m/s, while interpenetration volume (IV) and depth (ID) measure body--object overlap in $\mathrm{cm}^3$ and the maximum penetration depth in mm, respectively. Contact ratio (CR) denotes the proportion of body vertices within 5 mm of the object surface. We also report the average warm inference latency per sample (Time) in seconds.

\paragraph{Baselines.}
We compare with Dex2HOI~\citep{pratikaki2026dex2hoi}, an interaction-
representation baseline; DiffGrasp~\citep{zhang2025diffgrasp}, adapted from
object-conditioned diffusion to text-to-HOI; HOIDiNi~\citep{ron2025hoidini},
which jointly models human, object, and interaction states; and
CoDA~\citep{pi2026coda}, which uses staged generation and DNO. We report both
unrefined and guided/optimized variants when available. To isolate the effect
of the transport representation, we additionally train one-step MeanFlow
models in the Dex2HOI and HOIDiNi state spaces. Implementation details are provided in the Supplementary Material.

\paragraph{Qualitative Comparison.}

\begin{table}[t]
  \centering
  \fontsize{9}{11}\selectfont
  \rmfamily
  \setlength{\tabcolsep}{1mm}
  \begin{tabular}{@{}lrrrrrrr@{}}
    \toprule
    Method
    & FID$\downarrow$
    & R@3$\uparrow$
    & Div.$\rightarrow$
    & IV$\downarrow$
    & ID$\downarrow$
    & CR$\uparrow$
    & Time$\downarrow$ \\
    \midrule
    GT
    & 0.00 & 0.55 & 13.24
    & 3.97 & 3.49 & 0.64 & -- \\
    Dex2HOI
    & 3.44 & 0.30 & 10.15
    & 5.14 & 1.97 & 0.33 & 0.19 \\
    DiffGrasp-U
    & 5.15 & 0.27 & 10.62
    & 5.93 & 5.29 & 0.38 & 1.63 \\
    DiffGrasp-G
    & 4.26 & 0.31 & 10.70
    & 5.96 & 5.75 & 0.40 & 363.47 \\
    HOIDiNi
    & 5.81 & 0.34 & 11.47
    & 4.14 & 1.21 & 0.06 & 0.19 \\
    HOIDiNi(DNO)
    & 5.42 & 0.25 & 11.55
    & 4.79 & 4.47 & 0.64 & 379.50 \\
    CoDA (DNO)
    & 7.52 & 0.41 & 9.77
    & 4.82 & 5.20 & 0.57 & 872.95 \\
    \midrule
    \method{}
    & 2.57 & 0.45 & 11.25
    & 3.58 & 1.45 & 0.09 & 0.01 \\
    + Obj. Ref.
    & 2.22 & 0.42 & 11.23
    & 4.58 & 5.21 & 0.53 & 1.82 \\
    + UB-DNO
    & 7.25 & 0.35 & 9.56
    & 5.92 & 6.68 & 0.69 & 16.54 \\
    \bottomrule
  \end{tabular}
  \caption{Comparison on GRAB. U and G denote unguided and guided generation, respectively;
Obj. Ref. denotes object refinement, and UB-DNO denotes upper-body DNO.}
  \label{tab:main}
\end{table}

\begin{figure}[t]
  \centering
  \includegraphics[width=\columnwidth]{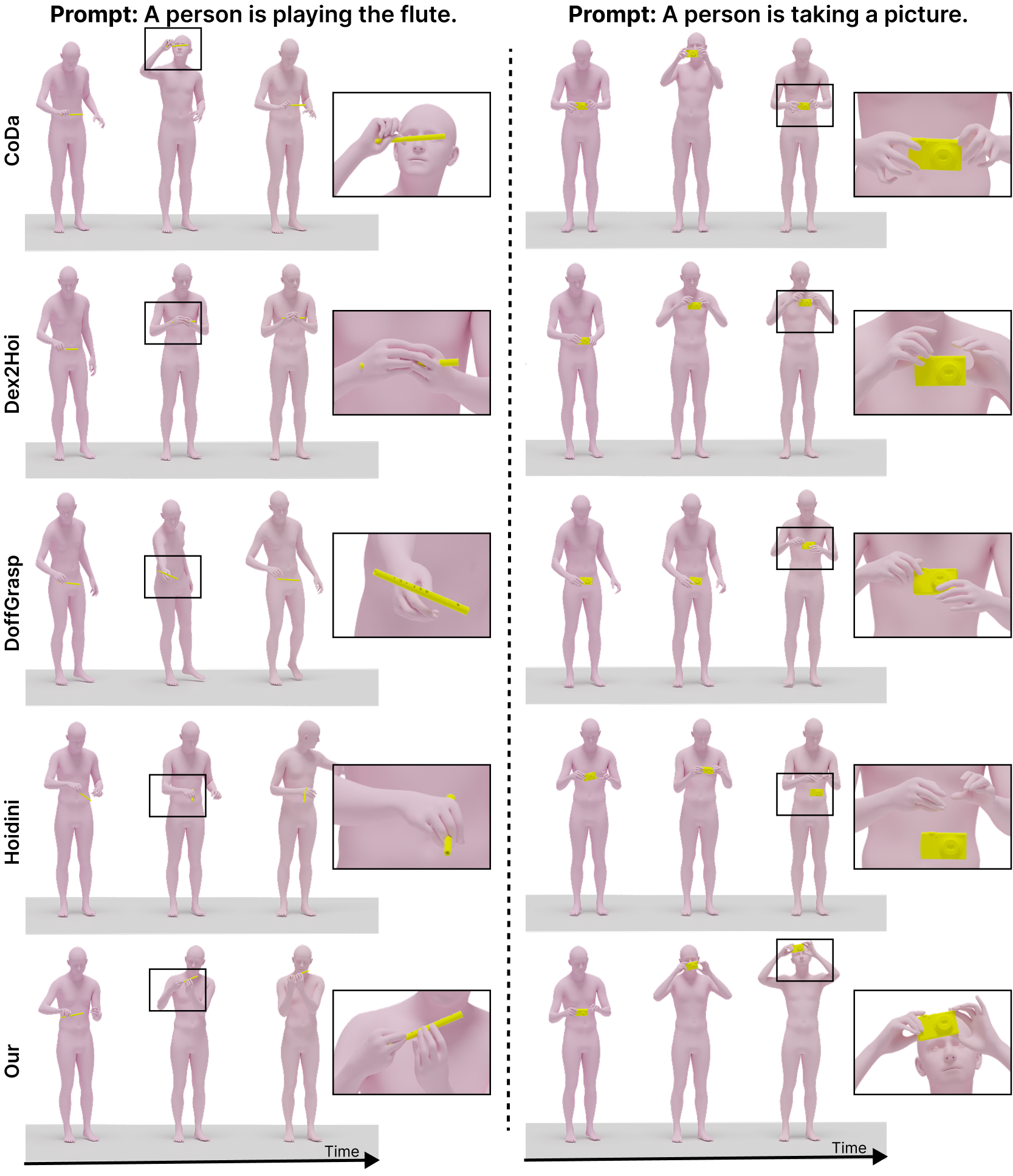}
  \caption{Qualitative comparison results. Columns show two prompts, and rows
compare different methods using matched frames and hand--object close-ups.}
  \label{fig:motion_comparison}
\end{figure}

\begin{table}[t]
  \centering
  \small
  \setlength{\tabcolsep}{2.0pt}
  \begin{tabular}{@{}lrrrrr@{}}
    \toprule
    Method
    & FID$\downarrow$
    & R@3$\uparrow$
    & Div.
    & FS$\downarrow$
    & CR$\uparrow$ \\
    \midrule
    Dex2HOI space
    & 7.2769 & 0.3359 & 9.8175 & 1.0859 & 0.0815 \\
    HOIDiNi space
    & 13.1630 & 0.2109 & 9.2114 & 3.4212 & 0.0300 \\
    \method{} w/o ST Attn.
    & 3.1960 & 0.3828 & 11.9047 & 0.1021 & 0.0524 \\
    \method{}
    & \textbf{2.5732}
    & \textbf{0.4531}
    & 11.2474
    & \textbf{0.0929}
    & \textbf{0.0930} \\
    \bottomrule
  \end{tabular}
  \caption{One-step MeanFlow results using different transport
  representations. All variants perform one full-interval transport
  evaluation without DNO. FS denotes foot skating.}
  \label{tab:meanflow_ablation}
\end{table}

\begin{table}[!b]
  \centering
  \small
  \setlength{\tabcolsep}{2.0pt}
  \begin{tabular}{@{}lrrrr@{}}
    \toprule
    Method
    & R-MPJPE$\downarrow$
    & FID$\downarrow$
    & CP-F1$\uparrow$
    & CR$\uparrow$ \\
    \midrule
    GT
    & 0.0 & 0.000 & -- & 0.6387 \\
    Holistic (w/o Int.)
    & 139.4 & 3.830 & -- & 0.1409 \\
    Part-wise (w/o Int.)
    & 47.3 & 0.190 & -- & 0.2852 \\
    \midrule
    Ours (Part-wise; w Int.)
    & \textbf{33.5}
    & 0.133
    & \textbf{0.956}
    & 0.1345 \\
    Ours + Obj. Ref.
    & \textbf{33.5}
    & \textbf{0.065}
    & \textbf{0.956}
    & \textbf{0.6432} \\
    \bottomrule
  \end{tabular}
  \caption{Latent reconstruction results. R-MPJPE measures root-relative joint reconstruction error, while CP-F1 evaluates reconstructed contact probabilities $\gamma$.}
  \label{tab:latent_ablation}
\end{table}

\begin{table*}[!t]
  \centering
  \small
  \setlength{\tabcolsep}{1.4pt}
  \begin{tabular}{@{}lrrrrrrrr@{}}
    \toprule
    & \multicolumn{3}{c}{Objective evaluation}
    & \multicolumn{5}{c}{Perceptual study} \\
    \cmidrule(lr){2-4}\cmidrule(l){5-9}
    Method
    & OBJ-CLIP$\uparrow$
    & LMD-B$\downarrow$
    & LMD-H$\downarrow$
    & Realism$\uparrow$
    & App.-H$\uparrow$
    & App.-O$\uparrow$
    & Mot.-H$\uparrow$
    & Mot.-O$\uparrow$ \\
    \midrule
    Animate Anyone
    & 0.6693 & 13.98 & 39.33
    & 1.46 & 2.03 & 1.24 & 2.37 & 1.25 \\
    Ours (3DGS)
    & 0.8067 & 11.05 & 21.87
    & 4.35 & 4.51 & 4.49 & 4.70 & 4.53 \\
    \bottomrule
  \end{tabular}
  \caption{Comparison with Animate Anyone. OBJ-CLIP measures object appearance
  similarity; LMD-B/H are body/hand landmark errors in pixels. Perceptual
  scores are averaged over 16 participants on a 1--5 scale and assess overall
  realism, appearance consistency (App.), and motion smoothness (Mot.) for the
  human (H) and object (O). Each method uses its respective 2D or 3D assets.}
  \label{tab:rendering_pipeline}
\end{table*}

Table~\ref{tab:main} compares generation quality, interaction geometry, and
runtime on GRAB. Without test-time optimization, \method{} achieves the best
FID (2.57) and R@3 (0.45) among the evaluated non-reference base
generators, while requiring only 0.0142 seconds per sequence. Similar to
HOIDiNi without DNO, the direct model has a relatively low CR, indicating that
semantic motion generation alone does not guarantee exact surface contact.
Unlike Dex2HOI, however, \method{} explicitly models bidirectional
human--object contact relations, which provide actionable constraints for
subsequent object- or body-side correction. We therefore use the base model to
prioritize efficient and semantically faithful generation, while treating
geometric refinement as an optional stage.
Existing refinement mechanisms improve contact-related metrics but require
expensive iterative inference: DiffGrasp with reconstruction guidance takes
363.47 seconds, while the DNO variants of HOIDiNi and CoDA require 379.50 and
872.95 seconds, respectively. In contrast, our predicted contact relations
enable lightweight bidirectional refinement. Object refinement increases CR
from 0.0930 to 0.5323 in 1.8167 seconds, while 20-step upper-body DNO reaches
0.6936 in 16.5422 seconds. The latter degrades FID and penetration metrics,
confirming that stronger contact does not necessarily imply better overall
generation quality. These variants provide complementary operating points:
direct generation prioritizes quality and efficiency, object refinement
provides balanced contact correction, and upper-body DNO targets
contact-critical applications. 

Figure~\ref{fig:motion_comparison} further illustrates these semantic and
geometric differences. \method{} preserves
action-defining spatial relations, such as raising the flute toward the mouth
with a coordinated two-handed configuration and positioning the camera near
eye level. Competing methods may generate plausible gross body motion while
placing the object away from these action-specific target regions. The
close-ups also show temporally coordinated hand--object configurations across
the selected frames, rather than contact that occurs only at isolated
instants. These observations are consistent with the improved text--motion
alignment in Table~\ref{tab:main}, while the remaining small contact gaps in
the direct outputs motivate the optional geometric refinement.

\subsection{Ablation Study}

\paragraph{Structured HOI Representation.}
As shown in Table~\ref{tab:latent_ablation}, part-wise encoding reduces
R-MPJPE from 139.4 to 47.3\,mm and FID from 3.830 to 0.190 compared with
holistic encoding. Separating body, hand, and object signals prevents
large-scale body motion from dominating finer interaction dynamics.
Adding explicit interaction states further reduces R-MPJPE to 33.5\,mm and
FID to 0.133, while achieving a C-F1 of 0.956. Object refinement improves CR
from 0.1345 to 0.6432 and FID from 0.133 to 0.065 without affecting human
motion. This indicates that the predicted interaction states provide
actionable geometric correspondences for correcting object placement.
\begin{figure}[h]
  \centering
  \includegraphics[width=\columnwidth]{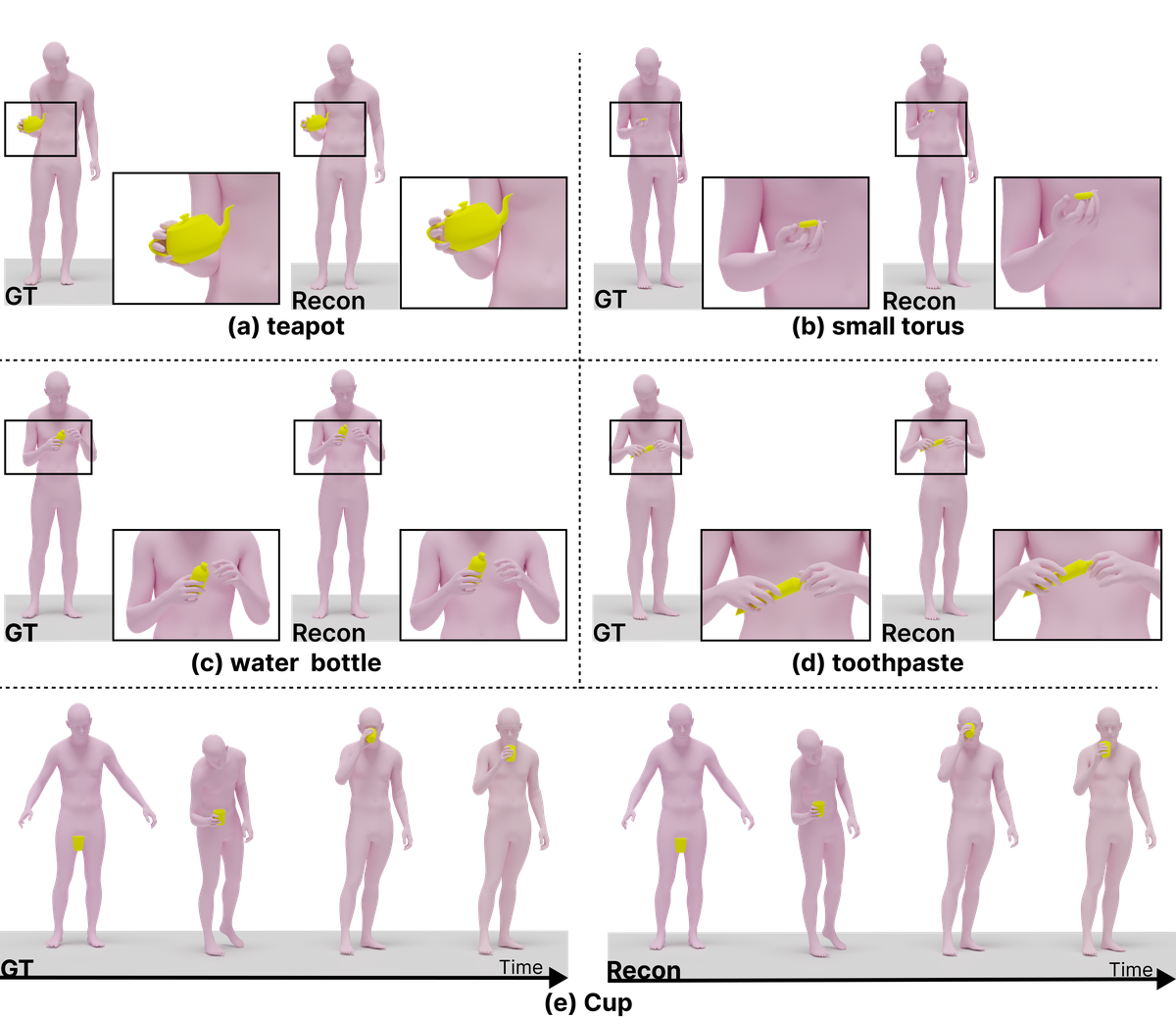}
  \caption{Reconstruction results. Panels (a)--(d) compare
  ground truth with reconstructed frames and hand--object close-ups; panel (e) compares the interaction sequences.}
  \label{fig:latent_ablation_qualitative}
\end{figure}
Figure~\ref{fig:latent_ablation_qualitative} shows that our codec preserves object placement, hand configurations, and their temporal evolution across different objects and interaction patterns. This confirms the structured representation captures coordinated human–object motion rather than only body poses.

\paragraph{Structured Latent Space for One-step Generation.}
As reported in Table~\ref{tab:meanflow_ablation}, directly applying MeanFlow
in the Dex2HOI and HOIDiNi spaces yields FIDs of 7.2769 and 13.1630,
respectively, with severe foot skating. In contrast, our structured latent
space without ST attention reduces FID to 3.1960 and FS to 0.1021.
These results show compressing heterogeneous HOI signals into balanced latent streams provides a better-conditioned space for one-step transport.

\paragraph{Spatiotemporal Interaction Modeling.}
ST-DiT further reduces FID from 3.1960 to 2.5732 and FS from 0.1021 to
0.0929, while improving R@3 from 0.3828 to 0.4531 and CR from 0.0524 to
0.0930. By modeling temporal dependencies within each stream and interactions
across streams, ST-DiT improves semantic alignment, motion quality, and
physical plausibility.

\subsection{Photorealistic 3DGS Deployment}
\begin{figure}[ht]
  \centering
  \includegraphics[width=\columnwidth]{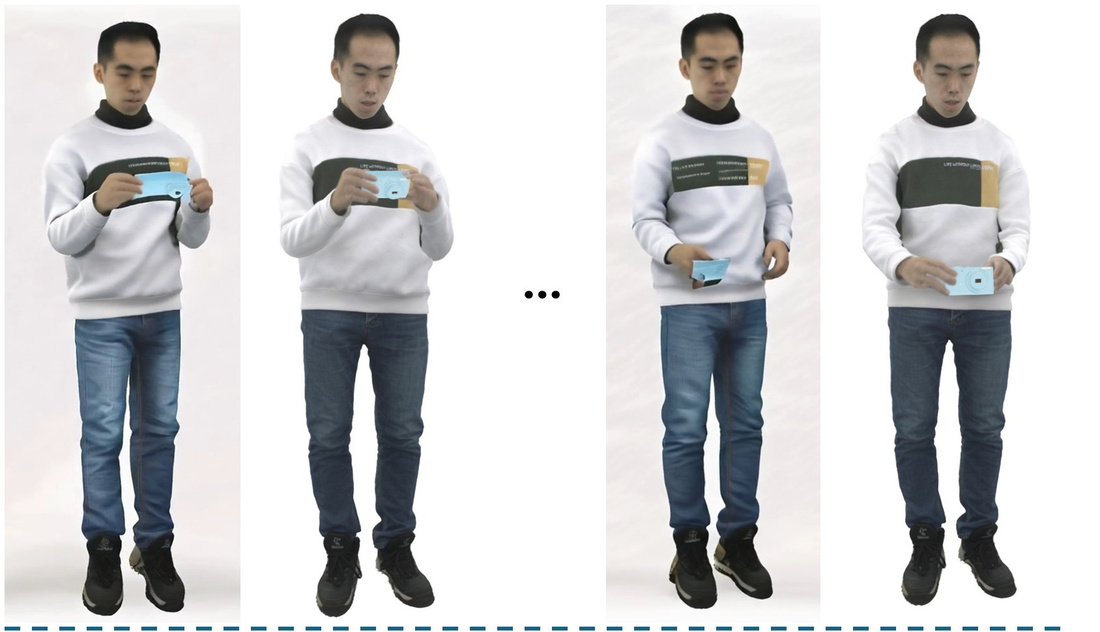}
  \caption{Qualitative comparison of Animate Anyone (first and third columns) and our
3DGS pipeline (second and fourth columns) at two matched instants under their
respective asset regimes. Ours better preserves object identity, temporal
coherence, and hand--object alignment.}
  \label{fig:rendering_comparison}
\end{figure}

We compare our 3DGS-based rendering pipeline with the 2D pose-driven Animate
Anyone~\citep{hu2024animate} under their respective asset regimes.
As reported in Table~\ref{tab:rendering_pipeline}, our approach improves
OBJ-CLIP from 0.6693 to 0.8067 and reduces body and hand landmark errors from
13.98 to 11.05 and from 39.33 to 21.87, respectively. It also achieves higher
perceptual scores across all evaluation criteria, with particularly large
gains in realism (4.35 vs.\ 1.46), object appearance consistency
(4.49 vs.\ 1.24), and object motion smoothness (4.53 vs.\ 1.25).
The qualitative comparisons in Figure~\ref{fig:rendering_comparison} further
show that our explicit 3D representation better preserves object identity,
temporal coherence, and hand--object alignment. Additional results evaluating
cross-shape interaction preservation are provided in the supplementary
material.

\section{Conclusion and Limitations}
\label{sec:conclusion}
We presented \method{}, a one-step framework for whole-body HOI generation. Its contact-aware multipart codec balances body, hand, object, and contact signals while preserving global coordination. Operating on the resulting time--part latent grid, ST-DiT generates complete interactions with a single MeanFlow evaluation. Experiments demonstrate improved distributional quality and text alignment compared with other generators, at substantially lower latency. Predicted object-local contacts further facilitate efficient geometric correction and avatar deployment. Although promising results were achieved, our evaluation is limited to interactions with a single rigid object. Extending the method to articulated/deformable objects remains future work.

\bibliography{aaai2027}

\end{document}